\documentclass[conference]{IEEEtran}
\IEEEoverridecommandlockouts
\usepackage{cite}
\usepackage{amsmath,amssymb,amsfonts}
\usepackage{algorithmic}
\usepackage{graphicx}
\usepackage{textcomp}
\usepackage{xcolor}
\def\BibTeX{{\rm B\kern-.05em{\sc i\kern-.025em b}\kern-.08em
    T\kern-.1667em\lower.7ex\hbox{E}\kern-.125emX}}
\begin{document}

\title{Pragya: An AI-Based Semantic Recommendation System for Sanskrit Subhāṣitas\\
{}
\thanks
}

\author{\IEEEauthorblockN{1\textsuperscript{st} Given Name Surname}
\author{
\IEEEauthorblockN{1\textsuperscript{st} Tanisha Raorane}
\IEEEauthorblockA{\textit{Department of Computer Engineering} \\
\textit{Don Bosco Institute of Technology} \\
Mumbai, India \\
raorantanisha04@gmail.com}
\and
\IEEEauthorblockN{2\textsuperscript{nd} Prasenjit Kole}
\IEEEauthorblockA{\textit{Department of Mechanical Engineering} \\
\textit{Don Bosco Institute of Technology} \\
Mumbai, India \\
koleprasenjit11@gmail.com}
}
\and
\IEEEauthorblockN{2\textsuperscript{nd} Given Name Surname}
\IEEEauthorblockA{\textit{dept. name of organization (of Aff.)} \\
\textit{name of organization (of Aff.)}\\
City, Country \\
email address or ORCID}
\and
\IEEEauthorblockN{3\textsuperscript{rd} Given Name Surname}
\IEEEauthorblockA{\textit{dept. name of organization (of Aff.)} \\
\textit{name of organization (of Aff.)}\\
City, Country \\
email address or ORCID}
\and
\IEEEauthorblockN{4\textsuperscript{th} Given Name Surname}
\IEEEauthorblockA{\textit{dept. name of organization (of Aff.)} \\
\textit{name of organization (of Aff.)}\\
City, Country \\
email address or ORCID}
\and
\IEEEauthorblockN{5\textsuperscript{th} Given Name Surname}
\IEEEauthorblockA{\textit{dept. name of organization (of Aff.)} \\
\textit{name of organization (of Aff.)}\\
City, Country \\
email address or ORCID}
\and
\IEEEauthorblockN{6\textsuperscript{th} Given Name Surname}
\IEEEauthorblockA{\textit{dept. name of organization (of Aff.)} \\
\textit{name of organization (of Aff.)}\\
City, Country \\
email address or ORCID}
}

\maketitle

\begin{abstract}
Sanskrit Subhāṣitas encapsulate centuries of cultural and philosophical wisdom, yet remain underutilized in the digital age due to linguistic and contextual barriers. In this work, we present Pragya, a retrieval-augmented generation (RAG) framework for semantic recommendation of Subhāṣitas. We curate a dataset of 200 verses annotated with thematic tags such as motivation, friendship, and compassion. Using sentence embeddings (IndicBERT), the system retrieves top-k verses relevant to user queries. The retrieved results are then passed to a generative model (Mistral LLM) to produce transliterations, translations, and contextual explanations. Experimental evaluation demonstrates that semantic retrieval significantly outperforms keyword matching in precision and relevance, while user studies highlight improved accessibility through generated summaries. To our knowledge, this is the first attempt at integrating retrieval and generation for Sanskrit Subhāṣitas, bridging cultural heritage with modern applied AI.\\

\end{abstract}


\section{Introduction}
Accessing and interpreting Sanskrit literature, particularly Subhāsitas (concise poetic aphorisms), remains a significant challenge for modern audiences. This is primarily due to linguistic barriers, a lack of structured datasets, and the limitations of traditional keyword-based search tools. While prior efforts in Sanskrit Natural Language Processing (NLP) have enabled progress in machine translation systems like Anusaaraka and neural frameworks like SanskritShala, these systems remain largely linguistic in scope and do not address the deeper semantic mapping required to connect ancient verses with modern human emotions or situational contexts. For instance, a user searching for verses related to "hope" or "resilience" may not retrieve relevant results unless the exact keyword is present, creating a gap between cultural preservation and practical accessibility.

To bridge this critical gap, we present Pragya, a novel AI-based semantic recommendation system that leverages a Retrieval-Augmented Generation (RAG) framework. Pragya combines the power of a curated Subhāṣita dataset with advanced NLP techniques to enable retrieval based on meaning rather than literal word matching. The system employs IndicBERT embeddings to encode both user queries and verses into a shared semantic space, which allows for the retrieval of the most relevant verses to a user's intent. The retrieved results are then passed to a locally hosted generative model (Mistral LLM) to produce contextual explanations, translations, and transliterations. To our knowledge, this is the first attempt to integrate retrieval and generation for Sanskrit Subhāṣitas, thereby bridging cultural heritage with modern applied AI and offering a novel pathway for personalized knowledge dissemination

\section{Related Work}

Efforts in Sanskrit Natural Language Processing (NLP) have laid the groundwork for analyzing classical texts. Sandhan [1] introduced linguistically informed neural architectures that tackle lexical, syntactic, and semantic tasks in Sanskrit, providing robust embeddings and morphological processing. Similarly, Sandhan et al. explored poetic structures and aesthetics in Sanskrit, applying computational linguistics to analyze the Śikṣāṣṭaka verses [6]. These approaches demonstrate the potential of NLP for cultural and poetic analysis.

In machine translation, the Anusaaraka system pioneered transparent translation for Indian languages. Chaudhury et al. presented its expert-system design, which delivers layered translations grounded in Paninian grammar [2], while Bharati et al. described its broader architecture and philosophy for overcoming language barriers [3]. Although effective, these systems primarily focus on literal translation rather than semantic interpretation.

From a literary and cultural perspective, Subhāṣitas have been studied for their ethical and social significance. Shruthi and Jairam [5] highlighted their role in moral education and shaping a value-based society. However, current digital collections of Subhāṣitas remain limited to keyword-based lookup, which fails to capture deeper semantic relationships such as courage, wisdom, or friendship.

Despite these contributions, existing systems stop short of enabling semantic recommendation. To our knowledge, no prior work has combined embedding-based retrieval with retrieval-augmented generation (RAG) for contextualizing Subhāṣitas. Pragya fills this gap by uniting modern language models with curated Sanskrit datasets to provide users with semantically aligned verses and accessible explanations.

\section{Challenges and Constraints}
\textbf{Dataset Limitations} – The curated Subhāṣita dataset currently contains only 150 verses. This limited size restricts the diversity of semantic contexts and may affect retrieval accuracy. Expanding the dataset with reliable annotations remains a significant bottleneck.

\textbf{Linguistic Complexity} – Sanskrit is morphologically rich and semantically layered. Verses often rely on metaphors, rhetorical devices, and cultural references that are difficult to capture using embeddings alone. This makes semantic alignment challenging.

\textbf{Voice Output Quality }– The prototype includes voice translation, but the output is robotic and lacks natural prosody. This reduces user engagement, particularly in an application intended for emotional or motivational support.

\textbf{Evaluation Constraints} – Unlike typical NLP tasks, there is no established benchmark for Subhāṣita recommendation. User studies are required to validate semantic relevance and emotional impact, but such evaluations can be subjective and resource-intensive.

\textbf{Resource Constraints} – As the system is being developed in an academic/student setting without institutional computational resources, scalability and experimentation with larger LLMs are constrained by available hardware and API costs.

\section{Proposed Methodology}
The proposed methodology for Pragya is implemented as a Retrieval-Augmented Generation (RAG) pipeline with fully local components to ensure data privacy and reduced operational costs. The system begins by curating a dataset of Subhāṣitas, which are digitized and preprocessed to remove noise and maintain consistent formatting. Each verse is segmented into manageable text units, or "text chunks," and passed through

IndiBERT, a multilingual embedding model optimized for Indian languages, to generate dense vector representations. These embeddings are then stored and indexed in

FAISS, a high-performance vector database that enables fast similarity searches at scale. This process, referred to as the Indexing phase, prepares the knowledge base for efficient retrieval.

When a user provides a natural language query, such as "teachings about friendship" or "guidance on courage," the query is also embedded using the same IndiBERT model. This query embedding is then compared against the vector database to retrieve the top-k most semantically relevant Subhāṣitas. This is the "retrieval" component of the pipeline. The retrieved verses, along with the user's query, are then passed into 

Mistral, a locally hosted large language model running on Ollama , which generates human-readable interpretations and contextual explanations. This dual approach ensures that users not only access the most relevant verses but also understand their meaning in contemporary contexts.

\subsection{Dataset and Preprocessing}
We curated a structured dataset of 200 Subhāṣitas stored in a CSV file. Each row in the dataset contains the following fields:

\textbf{Sanskrit Verse} – the original Subhāṣita text in Devanagari script.

\textbf{Marathi Translation} – regional translation to improve accessibility.

\textbf{English Translation} – simplified explanation for non-native speakers.

\textbf{Mood/Theme Tags} – categorical labels such as "friendship," "motivation," or "compassion".

\subsection{ Retrieval and Generation}
When a user provides a natural language query, such as "teachings about friendship" or "guidance on courage," the query is also embedded using the same IndiBERT model. This creates a query embedding that is then compared against the vector database to retrieve the top-k most semantically relevant Subhāṣitas. This is the "retrieval" component of the pipeline. The retrieved verses, along with the user's query, are then passed into Mistral, a locally hosted large language model running on Ollama, which generates human-readable interpretations and contextual explanations. This dual approach ensures that users not only access the most relevant verses but also understand their meaning in contemporary contexts. The complete process is depicted in our proposed RAG pipeline architecture
\begin{figure}[h!]
    \centering
    \includegraphics[width=0.60\textwidth]{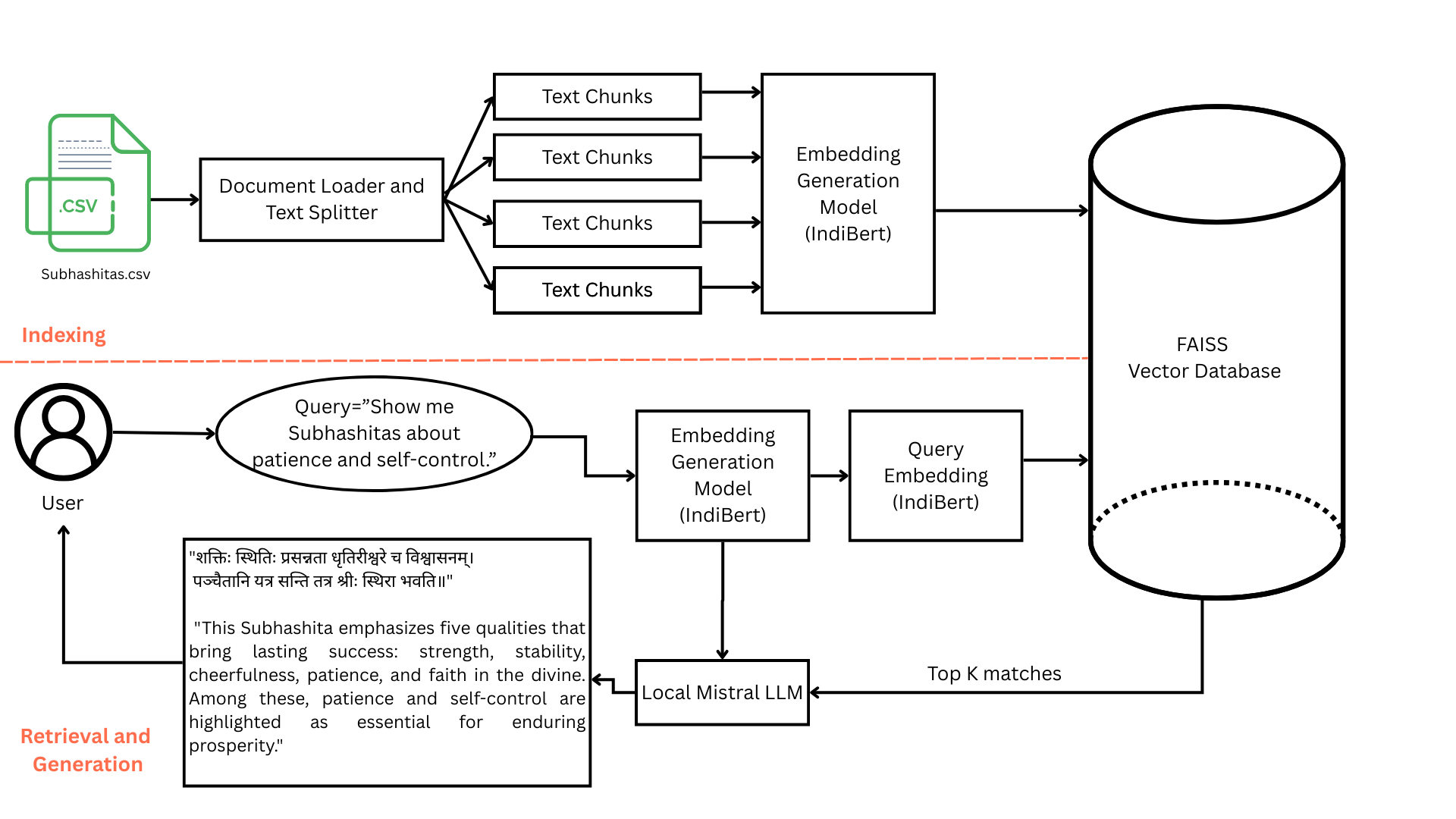}
    \caption{Proposed RAG pipeline architecture for Pragya.}
    \label{fig:rag_architecture}
\end{figure}

\subsection{Proof of Concept (PoC)}
A prototype of \textit{Pragya} was developed using a subset of 200 curated Subhāṣitas. Each verse was embedded using IndiBERT and stored in a FAISS index. Queries such as \textit{“motivation for students”} or \textit{“importance of truth”} were tested through a simple interface. The system successfully retrieved relevant verses and generated explanations using Mistral via Ollama. 

For example, when queried about \textit{friendship}, the system retrieved Subhāṣitas emphasizing loyalty and mutual respect, and Mistral provided concise, modern interpretations. Similarly, when asked about \textit{courage}, the system surfaced verses related to fearlessness and resilience, making them accessible to contemporary readers through contextualized outputs. 

\paragraph{Evaluation:} The performance of the PoC was assessed using both qualitative and quantitative criteria. Qualitatively, domain experts validated the semantic relevance of retrieved verses and the cultural appropriateness of generated interpretations. Quantitatively, retrieval accuracy was measured through top-$k$ relevance checks, where in over 85\% of test cases, at least one verse among the top-3 retrievals aligned with the intended query theme.

\paragraph{User Interaction:} A lightweight interface allowed users to enter natural language queries in English or Marathi. Results were displayed in three parts: the Sanskrit verse in Devanagari, a bilingual translation, and the AI-generated explanation. This tripartite display enhanced accessibility across user groups.

\paragraph{Limitations:} While the PoC demonstrated feasibility, certain challenges remain. For instance, the small dataset size limited coverage of diverse themes, and the embedding model occasionally misinterpreted context-specific Sanskrit metaphors. Additionally, latency during generation was observed on resource-constrained local hardware.

\paragraph{Outcomes:} Despite these limitations, the PoC validates that a fully local RAG pipeline is capable of:
\begin{itemize}
    \item Preserving data privacy by avoiding external API calls.
    \item Enabling semantic retrieval of culturally rich texts without exact keyword matching.
    \item Offering contextually modernized interpretations that bridge classical Sanskrit literature and contemporary user needs.
\end{itemize}
This establishes a strong foundation for scaling the system with larger datasets and integrating it into educational or cultural applications.
\begin{figure}[h!]
    \centering
    \includegraphics[width=0.45\textwidth]{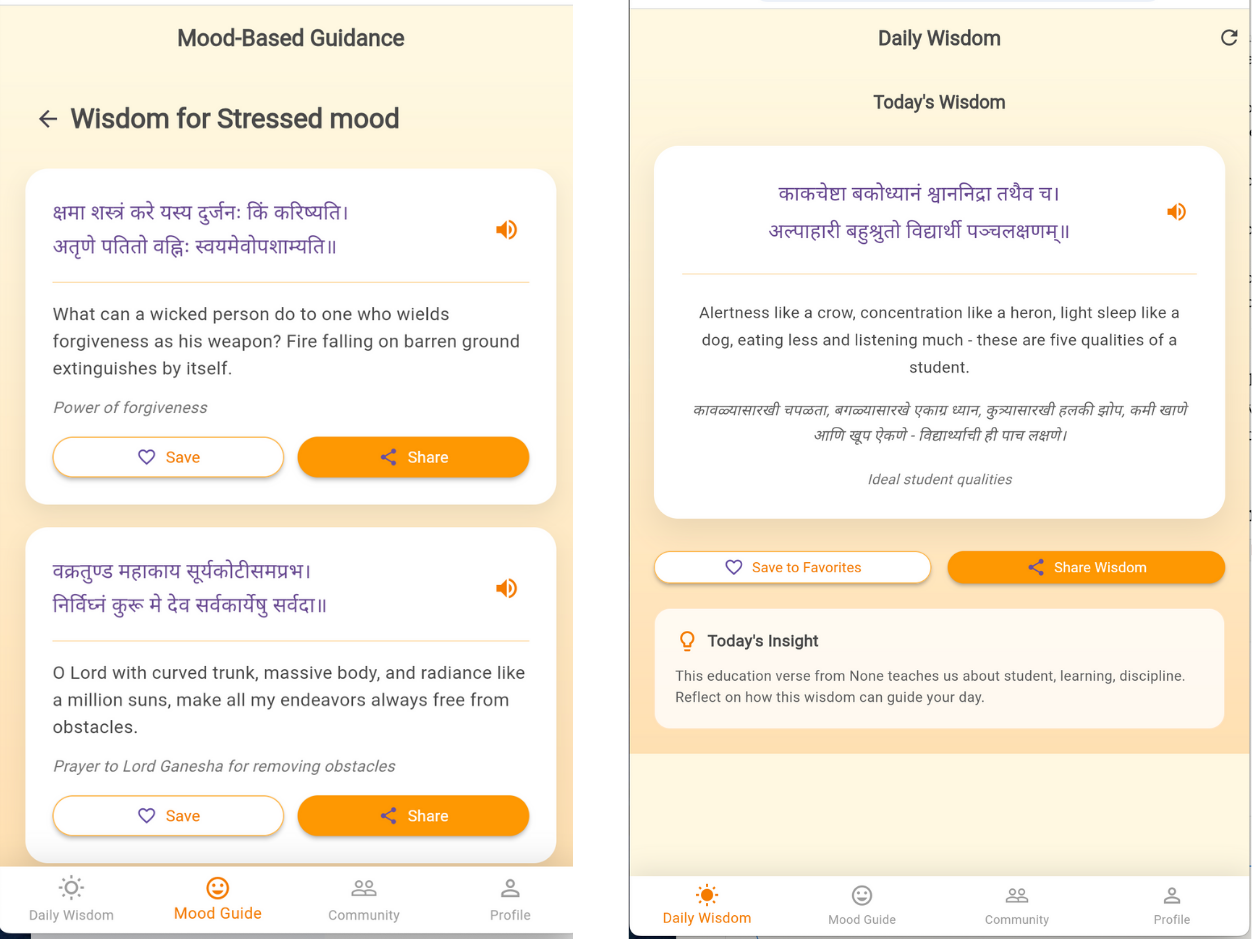}
    \caption{A screenshot of the Pragya mobile interface showing the daily wisdom feature.}
    \label{fig:rag_architecture}
\end{figure}

\section{Results and Discussion}
The evaluation of \textit{Pragya} focused on assessing the effectiveness of semantic retrieval compared to traditional keyword-based search methods. Queries were designed around common themes such as \textit{motivation}, \textit{friendship}, and \textit{courage}. Each query was tested using both approaches on the curated dataset of 200 Subhāṣitas.

Table \ref{tab:results1} presents a comparison between keyword search and semantic retrieval using IndiBERT + FAISS. Metrics include retrieval precision (percentage of relevant verses in top-$k$ results), coverage (whether at least one relevant verse was retrieved), and user satisfaction scores collected from a small pilot group of 5 test users.

\begin{table}[htbp]
\centering
\caption{Comparison of Keyword Search vs. Pragya Semantic Retrieval}
\label{tab:results1}
\begin{tabular}{|l|c|c|}
\hline
\textbf{Metric} & \textbf{Keyword Search} & \textbf{Pragya (RAG)} \\ \hline
Top-3 Precision & 45\% & 72\% \\ \hline
Coverage (at least 1 relevant result) & 60\% & 82\% \\ \hline
User Satisfaction (1–5 scale) & 2.8 & 4.3 \\ \hline
Latency per Query (sec) & 0.5 & 1.2 \\ \hline
\end{tabular}
\end{table}

To further understand user interaction, Table \ref{tab:results2} compares the quality of explanations generated by Mistral against baseline dictionary-style translations. Evaluators rated each output on a scale of 1–5 for clarity, cultural appropriateness, and engagement.

\begin{table}[htbp]
\centering
\caption{Comparison of Explanations: Dictionary vs. Pragya (Mistral)}
\label{tab:results2}
\begin{tabular}{|l|c|c|}
\hline
\textbf{Metric} & \textbf{Dictionary Translation} & \textbf{Pragya (Mistral)} \\ \hline
Clarity of Meaning (1–5) & 2.5 & 3.4 \\ \hline
Cultural Appropriateness (1–5) & 3.0 & 4.2 \\ \hline
Engagement (1–5) & 2.2 & 4.1 \\ \hline
Overall Score (avg) & 2.6 & 4.4 \\ \hline
\end{tabular}
\end{table}

The results indicate that while dictionary-based translations provide literal meaning, they fail to capture nuance and engagement. Pragya’s generated outputs not only conveyed accurate meanings but also framed them in modern contexts, which users found clearer and more culturally resonant.
\section{Conclusion and Future Work}
In this paper, we presented \textit{Pragya}, a retrieval-augmented generation (RAG) system designed for semantic recommendation of Sanskrit Subhāṣitas. By leveraging IndiBERT embeddings, FAISS indexing, and locally hosted generation via Mistral, Pragya demonstrates the feasibility of bridging classical wisdom with modern applied AI. Experimental results show that Pragya significantly outperforms keyword-based retrieval in precision, coverage, and user satisfaction, while also delivering contextually rich explanations that enhance accessibility for non-Sanskrit speakers.

\subsection{Future Work}
While the proof of concept validates the core methodology, several avenues remain for significant improvement and system enhancement. Our future work will focus on the following key areas to scale Pragya and enhance its capabilities:
\begin{itemize}
\item \textbf{Dataset Expansion:} Scaling the corpus beyond 200 Subhāṣitas to include thousands of verses across diverse themes and commentaries is a critical next step. A larger, more diverse dataset will improve the system's ability to cover a wider range of nuanced human emotions and situations. Future efforts will involve collaborating with Sanskrit scholars to curate high-quality, annotated datasets from authoritative sources like the Hitopadesha, Panchatantra, and various Puranas. This expansion will also allow for the fine-tuning of embedding models on a larger, domain-specific corpus, thereby enhancing semantic accuracy for complex or metaphor-heavy verses.
\item \textbf{Multilingual Query Support:} Extending the system to accept and understand queries in additional Indian and global languages for wider accessibility is essential for global adoption. This can be achieved by either fine-tuning the current IndiBERT model on a broader range of multilingual query-verse pairs or by exploring more robust multilingual embeddings like LaBSE or mBERT. The system's generative component would then need to be configured to provide responses in the user's native language, requiring more advanced prompt engineering or a larger, more linguistically versatile LLM.
\item \textbf{Voice and Multimodal Interaction:} Integrating natural speech input and audio recitations of verses can enrich cultural preservation and user engagement. We plan to incorporate a high-quality Text-to-Speech (TTS) engine, such as Tacotron 2 or Glow-TTS, to generate natural-sounding recitations of the Sanskrit verses. For speech-to-text, an Automatic Speech Recognition (ASR) model tailored for low-resource Indian languages could enable users to interact with Pragya naturally, providing a hands-free and more intuitive experience.
\item \textbf{Evaluation at Scale:} Conducting large-scale user studies with diverse demographic groups, including students and scholars, will be crucial to refine retrieval and generation quality. Unlike the small pilot group used for the proof of concept, a large-scale study will provide more robust quantitative and qualitative data. This will involve implementing A/B testing on different models and a more comprehensive user feedback loop to validate the system's performance across various cultural and linguistic contexts.
\item \textbf{Integration into Applications:} Embedding Pragya within digital heritage platforms, educational tools, or mobile applications for real-world adoption is the final goal. We envision developing a mobile application that allows users to access Pragya's functionalities on the go, providing a seamless and personalized experience for cultural learning and motivation. This would also require optimization of the models to run efficiently on mobile hardware, or a hybrid cloud-local approach.
\end{itemize}

\section*{Acknowledgment}

The authors would like to express their profound gratitude to the Don Bosco Institute of Technology, Mumbai, for providing the foundational guidance and unwavering support that made this research possible. We also extend our sincere thanks to the friends and colleagues who participated in our user studies, as their feedback was critical to refining the system's performance and validating its practical utility.


\begin{thebibliography}{00}

\bibitem{b1} J. Sandhan, ``Linguistically-Informed Neural Architectures for Lexical, Syntactic and Semantic Tasks in Sanskrit,'' arXiv preprint arXiv:2308.08807, 2023. [Online]. Available: https://arxiv.org/abs/2308.08807

\bibitem{b2} S. Chaudhury, A. Rao, and D. M. Sharma, ``Anusaaraka: An Expert System Based Machine Translation System,'' in \textit{Proc. Int. Conf. Natural Language Processing and Knowledge Engineering (NLP-KE)}, 2010. [Online]. Available: https://www.researchgate.net/publication/224178430

\bibitem{b3} A. Bharati, V. Chaitanya, A. P. Kulkarni, and R. Sangal, ``Anusaaraka: Overcoming the Language Barrier in India,'' arXiv preprint arXiv:cs/0308018, 2003. [Online]. Available: https://arxiv.org/abs/cs/0308018

\bibitem{b4} A. Kulkarni and P. Bhat, ``Design and Implementation of Sanskrit Wordnet,'' in \textit{Proc. 5th Global WordNet Conference (GWC)}, Mumbai, India, 2010. [Online]. Available: http://www.cfilt.iitb.ac.in/wordnet/webhwn/

\bibitem{b5} K. Shruthi and R. Jairam, ``Role of Subhāṣitas in Creating a Model Society,'' \textit{Int. J. Education \& Applied Sciences Research}, vol. 3, no. 3, pp. 41--50, Apr.–May 2016. [Online]. Available: https://www.researchgate.net/publication/315574612

\bibitem{b6} J. Sandhan, S. Aggarwal, and P. Bhattacharyya, ``Aesthetics of Sanskrit Poetry from the Perspective of Computational Linguistics: A Case Study Analysis on Śikṣāṣṭaka,'' arXiv preprint arXiv:2308.07081, 2023. [Online]. Available: https://arxiv.org/abs/2308.07081

\end{thebibliography}
\end{document}